%% file: main.tex
\documentclass{article}

\usepackage[preprint]{neurips_2026}

\usepackage[utf8]{inputenc}
\usepackage[T1]{fontenc}
\usepackage{hyperref}
\usepackage{url}
\usepackage{booktabs}
\usepackage{amsfonts}
\usepackage{amsmath}
\usepackage{nicefrac}
\usepackage{microtype}
\usepackage{graphicx}
\usepackage{subcaption}
\usepackage{multirow}
\usepackage{xcolor}
\usepackage{fvextra}
\usepackage[most]{tcolorbox}
\hypersetup{
  hidelinks,
  pdftitle={Where Knowledge and Authority Sit Changes What an Agent Benchmark Can Resolve},
  pdfauthor={Dan C. Hsu and Luke Lu},
  pdfkeywords={interactive agent evaluation, agent benchmarks, socially distributed information, multi-agent systems, tool-using agents, coordination}
}

\title{Where Knowledge and Authority Sit Changes\\What an Agent Benchmark Can Resolve}

\author{%
  Dan C. Hsu$^{1,2}$ \qquad
  Luke Lu$^{1}$\thanks{Corresponding author.} \\
  $^{1}$RedMind Research, San Francisco, CA, USA \\
  $^{2}$National Taiwan University, Taipei, Taiwan \\
  \texttt{\{dan,luke\}@redmindresearch.org}
}

\begin{document}

\maketitle

\begin{abstract}
Most agent benchmarks put facts, tools and permissions behind one interface.
Real organizations spread them across people. Incognita asks what happens when
the task and success criterion stay fixed but access does not. We transform
eighteen customer-service tasks into three settings: direct access, one known
intermediary, and six role-isolated participants whose capabilities must be
discovered. Across 864 trials with four models, social access reduced success
for every model; the pre-specified intervals excluded zero for two. The latest
tested model, gpt-5.6-sol, achieved the highest social-access success at 0.65, a
0.11 decrease from centralized indirect access with an interval that included
zero. Exploratory comparisons separated five of six model pairs under social
access, while neither centralized setting separated any at this sample size. In
post-hoc task-blocked tests, three pairwise interaction $p$-values remained
significant after multiplicity adjustment. A reference-relative reader
associates the wider gaps with failures to obtain needed information. Because
the data cannot distinguish ineffective requests by the evaluated agent from
inaccurate replies by simulated participants, the reader's labels describe
where trajectories stopped, not why models differed.
\end{abstract}

\section{Introduction}
\label{sec:intro}

Human societies exceed the knowledge and authority of any one person because
people act with partial knowledge and control, relying on role-holders who make
the rest reachable~\citep{schutz1946wellinformed,halpern1990commonknowledge}.
That dependence has a cost: groups systematically fail to surface uniquely held
information~\citep{stasser1985pooling}.

Most agent benchmarks instead provide one centralized agent-facing route to task
requirements~\citep{yao2024taubench,zhou2024webarena,trivedi2024appworld,lu2025toolsandbox}.
This makes completion measurable but leaves open whether competence transfers to
settings where knowledge and authority must be reached through other people.

On eighteen customer-service tasks held fixed across three conditions, the
pre-specified result is lower social-access success for all four models and
intervals excluding zero for two. Exploratorily, the centralized conditions
separate no pair, while social access separates five of six; a task-blocked,
multiplicity-adjusted difference-in-differences analysis supports three pairwise
interactions. Ceiling removal does not change the asymmetry. Under social access,
reference-relative missing-acquisition labels decline from 37 to 10 across the
model range while execution-stage labels remain nearly flat, without
distinguishing an agent request from a participant response.

We obtain these matched conditions with Incognita, an environment framework built
on Concordia~\citep{vezhnevets2023concordia} that transforms the tasks of an
existing benchmark by assigning their knowledge and action authority to
role-isolated entities while preserving the criterion for success, which turns a
background condition of human action into a controlled variable.

This paper contributes: (1) evidence, under one unchanged success criterion,
that the tested access-structure bundle changes pairwise model gaps, including an
exploratory interaction test and post-hoc corpus-size extrapolations; (2) process
measures and a five-type, trajectory-only failure taxonomy that associate model
gaps with acquisition-stage failures and refine the premature-termination mode
of existing multi-agent taxonomies~\citep{cemri2025mast}; and (3) an instrument
that converts an existing benchmark into matched access conditions, with a frozen
coordination reference and deterministic, model-free trajectory reader.

\section{Socially distributed task environments}
\label{sec:sdte}

A \emph{socially distributed task environment} is an interactive environment in
which task-relevant knowledge is partitioned across role-isolated participants,
and consequential actions are reachable only through interaction with those
participants. Interaction is then part of the causal structure of the task,
because it governs both what the agent can know and what it can cause.

The definition names two properties, although the present intervention changes
them together with other features of interaction. Knowledge is
partitioned when no participant holds all of it and the holder of a given fact is
not announced. Authority is partitioned when the agent cannot perform a
consequential action itself and must obtain it from whichever participant holds
it. The three conditions of Section~\ref{sec:incognita} are the three settings of
these properties while preserving task cards and the success criterion. Under Centralized (direct) the
agent holds all authority and consults only the customer. Under Centralized
(indirect) authority sits with one declared participant, so it is mediated but
not partitioned. Under Socially Distributed both properties hold.

This isolates a competence a centralized interface cannot exercise. An agent
operating a tool interface is judged on selecting and parameterizing operations.
An agent inside a social structure it does not control is judged in addition on
establishing who holds what, obtaining it, and directing the holder to act, which
is the management of dependencies between activities that the coordination
literature describes and whose first term the work on transactive memory and on
grounding
decomposes~\citep{malone1994coordination,wegner1987transactive,clark1991grounding}.
Our measurements follow that decomposition. Formally a trial is a partially
observable decision process whose action is a directed message and whose latent
state carries a fixed partition of knowledge, embedded in a partially observable
stochastic game with fixed other
players~\citep{kaelbling1998pomdp,hansen2004posg,bernstein2002decpomdp}. Across
conditions the observation function and the reachability of consequential actions
change, and the state space, the database transition and the reward do not.

Among the environments reviewed here, we found no matched benchmark
transformation that also withholds the map of who holds what. Benchmarks with
executable actions and verifiable outcomes give
one agent a centralized route to every tool and every
document~\citep{yao2024taubench,zhou2024webarena,trivedi2024appworld,lu2025toolsandbox}.
A dual-control successor to the source benchmark gives the customer tools of its
own and models the result as a decentralized decision
process~\citep{barres2025tau2}, which distributes authority between the agent and
one counterpart whose capabilities are known. Environments for
information-asymmetric collaboration and for social intelligence distribute
knowledge across participants but score answers or dialogue rather than an
irreversible change to shared
state~\citep{liu2024infoasymmetry,zhou2024sotopia,kim2023fantom}. Environments
that do combine distributed knowledge with consequential state place the agent
inside an organization whose structure it is told, such as a simulated company
with named colleagues and departments~\citep{xu2025theagentcompany}. Environments
that enforce role separation declare each role and its permissions in
advance~\citep{kim2026teambench}, and environments for distributed coordination
that do not name roles give every agent the same capabilities over a shard of one
computation~\citep{zhang2026silobench}, so the problem there is organizing
symmetric peers. Closest in phenomenon is work on hidden profiles, where
distributing the evidence across agents collapses collective reasoning relative
to giving one agent everything~\citep{li2026hiddenbench}. That work scores a joint
answer produced by symmetric participants. Here a single evaluated agent must
first establish which asymmetric participant is able to act, and is scored on an
irreversible change to shared state.

\section{Incognita}
\label{sec:incognita}

Incognita transforms the tasks of an existing benchmark into task environments
with a common success criterion and different access structures. It is built on
Concordia,
which supplies a substrate in which language agents act on state that a game
master arbitrates~\citep{vezhnevets2023concordia}. Incognita specializes that
substrate for evaluation by separating a social interaction layer, in which the
evaluated agent reaches other participants only by message, from a deterministic
execution layer, in which accepted operations change a canonical database.

\begin{figure}[t]
  \centering
  \includegraphics[width=0.92\linewidth]{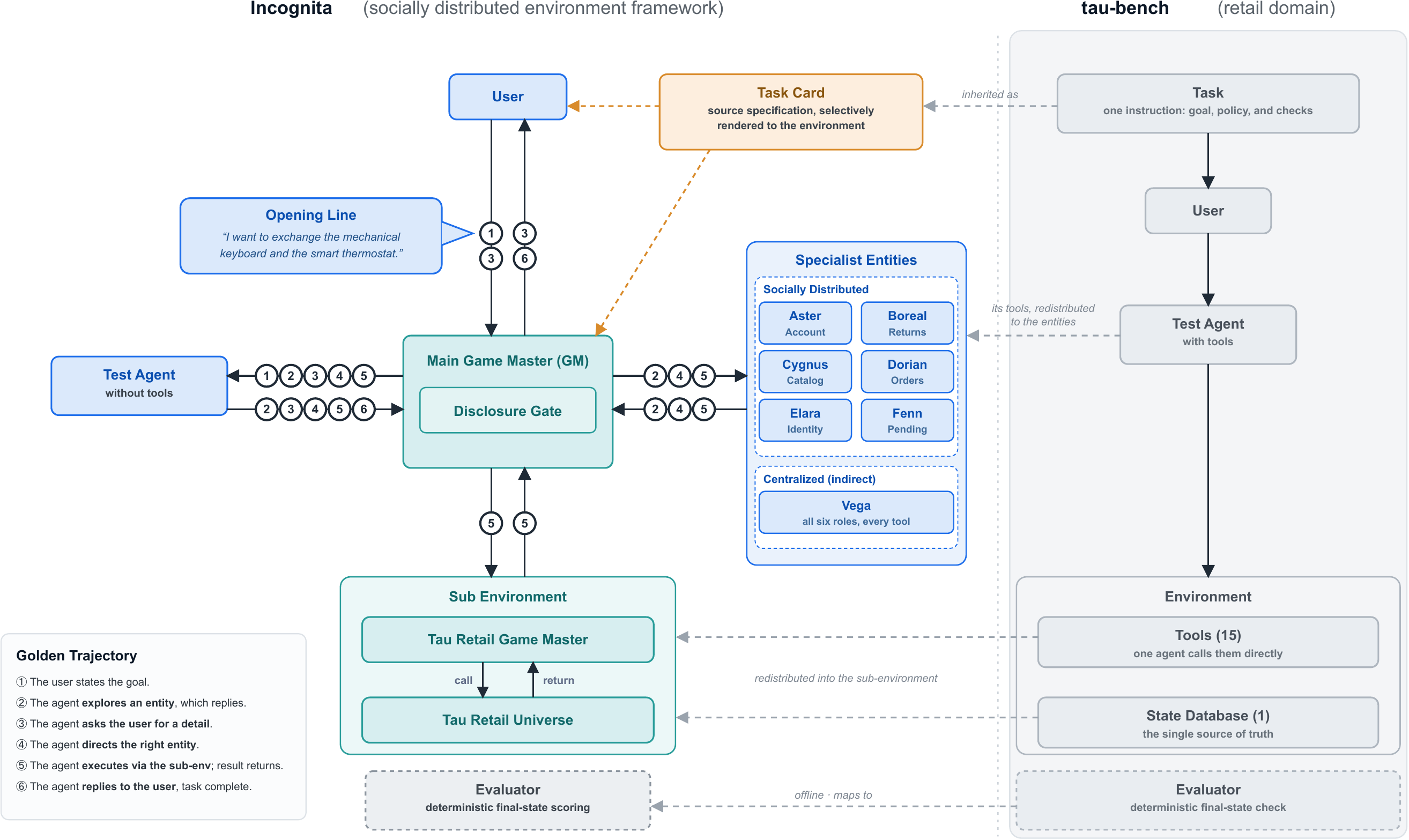}
  \caption{How a $\tau$-bench retail task becomes an Incognita task. The source
  benchmark, right, gives one agent the instruction, every tool and the state
  database. Incognita, left, keeps that specification as a task card and renders
  it selectively: the evaluated agent holds no tools and reaches the world only
  through routed messages, the disclosure gate releases the customer's facts only
  when asked, and the evaluator scores the final state exactly as the source does.
  The Specialist Entities block is the only element differing between the two
  indirect conditions, six role-isolated entities with undeclared tools under
  Socially Distributed against the single entity Vega under Centralized
  (indirect). Numbered badges trace the golden trajectory of the example task.}
  \label{fig:architecture}
\end{figure}

\subsection{The transformation}

The source specification of a task becomes a \emph{task card}. The card is not
shown to the evaluated agent. It is rendered into the environment selectively,
according to which participant holds each part of it and when that part becomes
available. Two axes carry the rendering. Demand-side knowledge, the facts that
only the customer knows, is held by a simulated USER behind a disclosure gate
that releases one field at a time and only when the agent asks for that field.
Service-side knowledge, what can be inspected or changed and by whom, is held by
the participants that mediate the tools. At initialization the agent observes the
participant list and one opening line from the customer, and nothing else.

The transformation leaves the source task's criterion for success untouched. The
evaluator reads the final database state and applies the inherited reward. This
is what makes the conditions below comparable: they change how the agent reaches
what the task requires, and they do not change what the task requires.

\subsection{Three access conditions and an external anchor}

Three conditions place the same eighteen tasks under different arrangements of
knowledge and authority. Every condition keeps the same task cards, the same
customer simulator and disclosure gate, the same termination contract, the same
evaluator and the same sampling parameters.

\paragraph{Centralized (direct).} The evaluated agent holds the union of the
service-side tools itself and calls them directly. The customer remains a
separate participant reached by message. No service-side entity exists. This is
the interface of the source benchmark, reproduced inside the Incognita harness.

\paragraph{Centralized (indirect).} A single entity, Vega, holds all service-side
knowledge and all fifteen tools. The evaluated agent holds no tools and reaches
every consequential action by messaging Vega. The capability of the one
participant is declared in its role description, so there is nothing to discover
about who holds what.

\paragraph{Socially Distributed.} The same knowledge and the same tools are
partitioned across six role-isolated entities. Each entity answers only within
its own remit, sees neither the customer dialogue nor the other entities, and
never refers the agent elsewhere. The entities are named Aster, Boreal, Cygnus,
Dorian, Elara and Fenn, and the assignment of names to capabilities is
decorrelated from any order the agent could guess. No entity announces its tools.
The agent must therefore establish who holds what by interacting, and then reach
the action through whichever participant holds it.

\paragraph{Native $\tau$-bench.} As an external anchor we run the same tasks and
models in the unmodified upstream harness at a pinned checkout, keeping its own
customer simulator, policy prompt and termination
rule~\citep{yao2024taubench}. It is not a fourth condition. It differs from
Centralized (direct) in several components at once, so the gap between them bounds
the cost of transporting the tasks into our harness and supports no causal claim.

The main comparison is the last step of this ladder, which holds the task cards,
the tool set, the customer, the termination contract, the evaluator and the
criterion for success fixed, and redistributes service-side knowledge and
authority. It is not a single-factor manipulation. Partitioning a service here
also changes the number of recipients, whether a capability map is declared,
role specialization, out-of-remit refusal, the absence of referral, and the
number of language-model participants. A misdirected question is therefore a
dead end rather than a redirection. Real organizations do refer. We treat this
full access-structure bundle as the manipulated variable and do not identify the
contribution of any one component.

\subsection{Grounded execution and evaluation}

The evaluated agent sends one directed message per turn, and sending a message
never ends the episode. The episode ends only on a separate termination signal to
a reserved recipient that never replies, which keeps a question to the customer
from ending the trial before the question is heard.

A participant that decides to act proposes a tool call. A deterministic
sub-environment checks that the proposer owns the tool and that the arguments are
admissible, then executes accepted calls against the canonical database. Writes
are irreversible, since the environment offers no operation that restores a
previous state.

Evaluation stays outside the interaction loop. After termination the evaluator
compares the final database state against a replayed gold state, and checks
natural-language assertions only for the seven tasks whose source specification
carries them. A trial succeeds when every operative reward channel passes.

\section{Coordination processes and failure types}
\label{sec:coordination}

Task success is a single bit and it does not say where a trajectory broke. This
section defines the measurements that do. Both are computed from structured
trajectory, gateway and disclosure events, without a model in the loop, and both
apply to all three conditions with the same definitions.

\subsection{The golden trajectory and the coordination reference}

For each task we build the complete sequence of operations that any condition
must perform to satisfy the criterion. The upstream reference actions are kept
verbatim as the write anchors, and the reads that those writes depend on are
added by rule, so that every argument of every call is traceable to a source: the
opening line, a field the disclosure gate can release, the customer's request, the
policy vocabulary, or the return value of an earlier call. Replaying the result on
a clean database reproduces the upstream final-state hash for all eighteen tasks.

Projecting this trajectory onto each condition changes only who executes each
call. That projection yields the \emph{coordination reference}: 146 calls, of
which 114 count toward the denominators below, 89 role assignments and 138
information items. Calls that the source task intends to fail, and reads that
serve only exploration, stay in the trajectory and stay out of the denominators.
The reference is frozen before any trajectory is read.

The reader does not infer facts from free-form natural-language messages. A
customer fact is acquired only when the disclosure gate records its structured
token as released. A service-side fact is acquired only when its designated
reference read is accepted with the reference arguments. Items whose required
value is not observable in these event records are excluded from the acquisition
denominator. The resulting reference is a diagnostic of dependencies needed by
one verified solution, not a claim that it is the only valid route to success:
the final-state evaluator is applied first, so a successful non-reference
trajectory is never labelled as a failure by this taxonomy.

\subsection{Three coordination processes}

Each process is scored per trial as the fraction of its required items that the
trajectory satisfies. A process that a condition does not structurally contain is
reported as absent and never as zero.

\begin{description}
\item[Role identification.] For each required call, whether the agent got the
participant who holds that tool to execute it. This process exists only where
holders are undeclared, so it is defined for Socially Distributed alone. We also
record the weaker event of contacting that participant at all, which separates
finding a holder from getting it to act.
\item[Task information acquisition.] Whether each fact a required call needs was
obtained: customer facts through a release by the disclosure gate, service facts
through a reference read that returned them.
\item[Task assignment and task execution.] Whether each required call was
executed by the expected executor with the reference arguments. Under indirect
access the agent assigns the call to a participant, and under direct access it
executes the call itself, so the same requirement is named assignment in the
first case and execution in the second.
\end{description}

Role identification and assignment are nested rather than independent, since
getting the right holder to execute a call is a precondition for that call
carrying the reference arguments. Their per-trial values are correspondingly
close under socially distributed access, and we report both because the gap
between them is small only when a located holder is also directed correctly. The
measures that carry weight below are information acquisition and the failure
types, which are defined on the same events in every condition.

\subsection{Five failure types}

We classify each failed trajectory by the stage of the coordination dependency
chain at which it had stopped when it terminated: exploration, in which the agent
obtains what the action needs, exploitation, in which it delivers the action, and
verification, in which the outcome is confirmed, the first two in the sense of
\citet{march1991exploration}. Each question is answered from trajectory events
and the first match is the label, so the types are mutually exclusive and
exhaustive over failed trials by construction. Where no action was attempted and
a required fact was never obtained, the label is \textbf{never requested} if the
agent never approached the source of any missing fact and \textbf{requested but
not obtained} if it did approach one without obtaining it. Where no action was
attempted but every required fact was obtained, the label is \textbf{obtained but
not delivered}. Where an action was attempted, the label is \textbf{delivered but
incorrect} if the required actions were not all executed with the reference
arguments or an action outside the reference was executed, and \textbf{correct
but unconfirmed} if they were and the task failed anyway.

The label \emph{requested but not obtained} records that a needed
reference-relative value remains absent after the relevant participant was
approached. It does not distinguish an inadequate request from an incorrect,
unsupported or refusing participant response. We therefore use these labels to
describe where a trajectory stopped, not to assign the cause of a failed
acquisition to the evaluated agent.

Types 1 to 3 are together the premature termination mode of the multi-agent
failure taxonomy of \citet{cemri2025mast}. That taxonomy records the mode as a
single category. The decomposition here supplies its internal structure, which is
what distinguishes an agent that never asked from one that asked and left without
an answer.

\section{Experimental setup}
\label{sec:setup}

\paragraph{Corpus.} Eighteen $\tau$-bench retail tasks, selected before any
Incognita run and frozen since, stratified by the number of role-isolated
entities a task's reference solution must engage, three at each level from one to
six, preferring tasks whose outcome includes a database write and excluding those
whose instructions withhold disclosure by design. The stratification is a
property of the socially distributed condition, so this corpus is not a random
sample of $\tau$-bench retail and nothing here estimates that benchmark's
behaviour.

\paragraph{Models and execution.} The evaluated agents are
gpt-5.2-2025-12-11, gpt-5.4-2026-03-05, gpt-5.5-2026-04-23 and gpt-5.6-sol, in
release order, at high reasoning effort and temperature 1, with three repetitions
per cell. The first three are dated snapshots; gpt-5.6-sol is the service
identifier returned by the API and had no immutable dated snapshot available to
us. Each of the three Incognita conditions and the native anchor contributes
216 trials, 864 in total. The Incognita customer simulator and every service-side
participant use gpt-5.5-2026-04-23 at temperature 0 and no reasoning effort. The
native customer simulator and the fixed natural-language judge use
gpt-4.1-2025-04-14. Calls were made on 2026-08-20--21, and costs use one
versioned price table retrieved on 2026-08-15. The agent is allowed 300 turns;
the longest trial in the whole run used 150, so the budget never binds and no
failure below is a truncation.

The natural-language channel is operative in 252 Incognita trials. In ten, the
database channel passes but the judge fails, so the judge determines the final
failure: one direct, four centralized-indirect and five socially distributed.

\paragraph{Schedule.} One worker runs the same task, model and repetition through
all four arms in sequence with the arm order rotated by a Latin square, which
keeps the arms of a comparison inside one time window.

\paragraph{Estimator.} Every quantity is aggregated at the task level. Within a
cell we average the repetitions of a task first, then average over the eighteen
tasks. Comparisons are per-task paired differences with a two-sided 95 percent
$t$ interval on the eighteen differences. We draw no endpoint error bars, because
task difficulty cancels inside a paired difference and does not cancel inside an
endpoint~\citep{miller2024errorbars}. Whether such an interval excludes zero is a
test at the 5 percent level, and where we count exclusions across the six model
pairs of an arm we say so and apply no multiplicity correction, so those counts
are descriptive comparisons between arms and not confirmatory findings.

\paragraph{Pre-specified analysis.} An internal interpretation table was frozen
on 2026-08-20 before the first paid call, after a 16-trial smoke test. It fixed
the primary comparison, the direction supporting the hypothesis, the floor and
ceiling checks, and the rule that an interval crossing zero may not be read as an
absence of effect. It was not externally deposited or publicly timestamped, so
we describe this analysis as \emph{pre-specified}, not pre-registered.
Section~\ref{sec:results} marks which results it specified and which diagnostics
came after the data were seen.

\paragraph{Task card repairs.} An audit after the first batch found eight of the
eighteen cards unfaithful to the source task, in every case on the customer side.
They were repaired under the rule that whatever the source tells the customer
must be releasable and whatever it does not tell the customer we do not invent.
The upstream reference actions were left untouched, and the 384 affected trials
were re-run in all four arms, the native anchor through a versioned data overlay.
The pre-repair data are archived and never pooled. The repairs were made after
the first batch, so we do not claim that they were selected without access to
outcomes; the pre-specified estimator and comparison were unchanged.

\section{Results}
\label{sec:results}

\subsection{Measured competence, and the resolution of the measurement}

\begin{figure}[t]
  \centering
  \includegraphics[width=\linewidth]{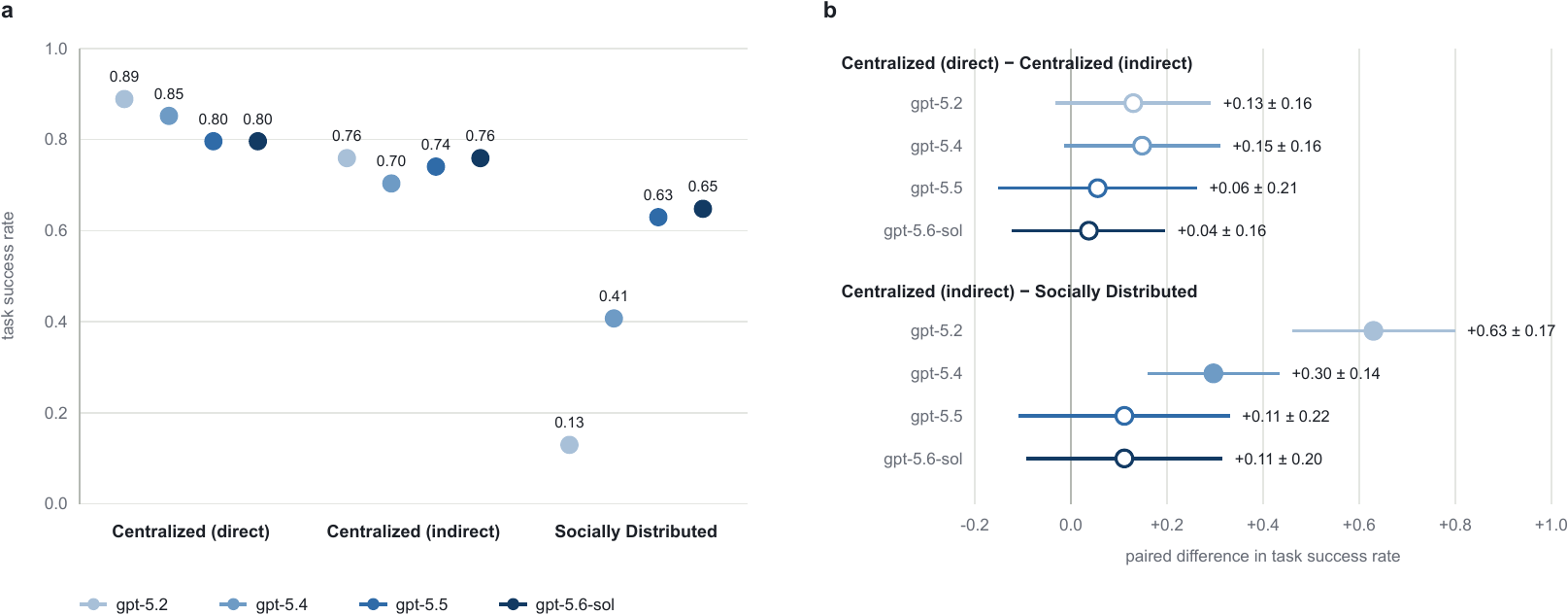}
  \caption{Task success across the three access conditions. (a) Mean task
  success rate of each model, with the rate printed above each dot. The last step
  is the main comparison, which redistributes knowledge and action authority.
  (b) Per-model paired differences between neighbouring conditions, computed per
  task with repetitions averaged first. The bar is a descriptive $t$ interval on
  the eighteen per-task differences and a filled dot marks an interval that
  excludes zero. Endpoint intervals are omitted because task difficulty cancels
  inside a paired difference and not inside an endpoint. The native anchor is
  compared with Centralized (direct) in Appendix Figure~\ref{fig:native}.}
  \label{fig:slope}
\end{figure}

Figure~\ref{fig:slope} gives the task success rate of every model in each
condition, together with the paired differences between neighbouring conditions.
Averaged over the four models, success falls from 0.83 under direct access to
0.74 under centralized indirect access and 0.45 under socially distributed
access, and the fall is carried very unevenly by the four models. The native
anchor reaches 0.85 and is shown in Appendix Figure~\ref{fig:native}.

\paragraph{The pre-specified comparison.} Moving from Centralized (indirect) to
Socially Distributed lowers success for all four models. We code this loss as
Centralized (indirect) minus Socially Distributed, so positive values mean lower
social-access success. The losses are $0.630 \pm 0.170$ for gpt-5.2 and $0.296
\pm 0.138$ for gpt-5.4, both excluding zero, and $0.111 \pm 0.220$ and $0.111
\pm 0.205$ for gpt-5.5 and gpt-5.6-sol, both including zero. The frozen table
therefore supports the hypothesis for the two weaker
models and finds the evidence insufficient for the two stronger, and forbids
reading an interval including zero as an absence of effect. No cell sits on a
floor or ceiling, though individual tasks do, which matters below.
Using database checks alone changes the estimates but not which two of the four
intervals exclude zero (Appendix~\ref{app:audit}).

The two supporting contrasts move less. Centralized (direct) exceeds Centralized
(indirect) by between $+0.037$ and $+0.148$, with all four intervals including
zero, so this design bounds the effect of routing every action through one
declared participant at roughly $0.15$ without establishing its size. The native
anchor and Centralized (direct) differ by between $-0.093$ and $+0.111$, again
with all four intervals including zero, which bounds the cost of transporting the
tasks into our harness.

\paragraph{An observation made after the data were seen.} The pre-specified result
is about how far scores fall. The more consequential pattern in the same data is
whether the scores separate the models at all, and we offer it as an exploratory
diagnostic. We also test it directly. After averaging repetitions within a task,
we swap the two indirect-condition labels within each task to form an exact
task-blocked sign-flip/permutation test under within-task label exchangeability
for each model-pair difference-in-differences.
The global increase in within-task between-model dispersion has two-sided
$p=0.0127$. Of the six pairwise interactions, the gpt-5.2 gaps from gpt-5.4,
gpt-5.5 and gpt-5.6-sol widen by $0.333$, $0.519$ and $0.519$, with Holm-adjusted
$p=0.0234$, $0.0037$ and $0.0132$. This post-hoc test supports an exploratory
condition-dependent pattern on this corpus, not a causal effect of one component
of the access-structure bundle. Treating each pair of models as a paired
comparison over the eighteen tasks, neither centralized condition nor the native
anchor separates a single pair of six. The best-to-worst spreads are 0.093 under
Centralized (direct), 0.056 under Centralized (indirect), and 0.111 under the
native anchor. Under socially distributed access five of six pairs separate, the
spread is 0.519, and only gpt-5.5 and gpt-5.6-sol remain inseparable at $-0.019
\pm 0.155$. Appendix Figure~\ref{fig:separability} shows every pair.

\emph{Zero of six is a bound, not an absence.} With eighteen tasks and the
observed dispersion of paired differences, the smallest effect this design
resolves at conventional power is about 0.20. Every centralized model-pair
difference is below that, so the correct reading is that no pair of models
differs by more than the design can see, not that the models are alike. The
corpus-size figures are post-hoc plug-in extrapolations: for each condition we
use the median observed standard deviation of its six task-level paired gaps and
its largest observed gap, target 80\% power at two-sided $\alpha=0.05$, and make
no multiplicity adjustment or uncertainty claim. On those assumptions, resolving
each condition's own largest observed model gap would take about 104 tasks under
Centralized (direct) and about 249 under Centralized (indirect), against about 6
under socially distributed access. The derivation and implied cost calculation are in
Appendix~\ref{app:interaction}.

Two simple rival accounts do not explain this pattern. Uniform difficulty would
lower every model and preserve the spread, whereas the drop here is 0.630 for one
model and 0.111 for another. Compression at the ceiling is the more serious
candidate, since a task every
model solves contributes an exact zero to every paired difference and the
centralized arms contain such tasks. Removing every task that all four models
solve or all four fail leaves 11 and 12 tasks in the centralized arms and 16
under socially distributed access, and the counts do not move: still zero of six
and five of six. We do not claim the recovered ordering is the correct one. A
thermometer that reads five degrees high still separates a warm day from a cold
one.

Reproducibility across halves of the task set is weaker than the separability
counts suggest. Splitting the eighteen tasks at random 4000 times, the socially
distributed condition agrees with itself on 5.0 of six pairwise orderings against
a null of 2.9, but 14.5 percent of null splits reach the observed value, which is
short of a conventional threshold. The three centralized measurements sit at or
below their nulls. We report this and do not lean on it. Appendix
Figure~\ref{fig:splithalf} shows the distributions.

\subsection{Where the trajectories break}

\begin{figure}[t]
  \centering
  \includegraphics[width=0.88\linewidth]{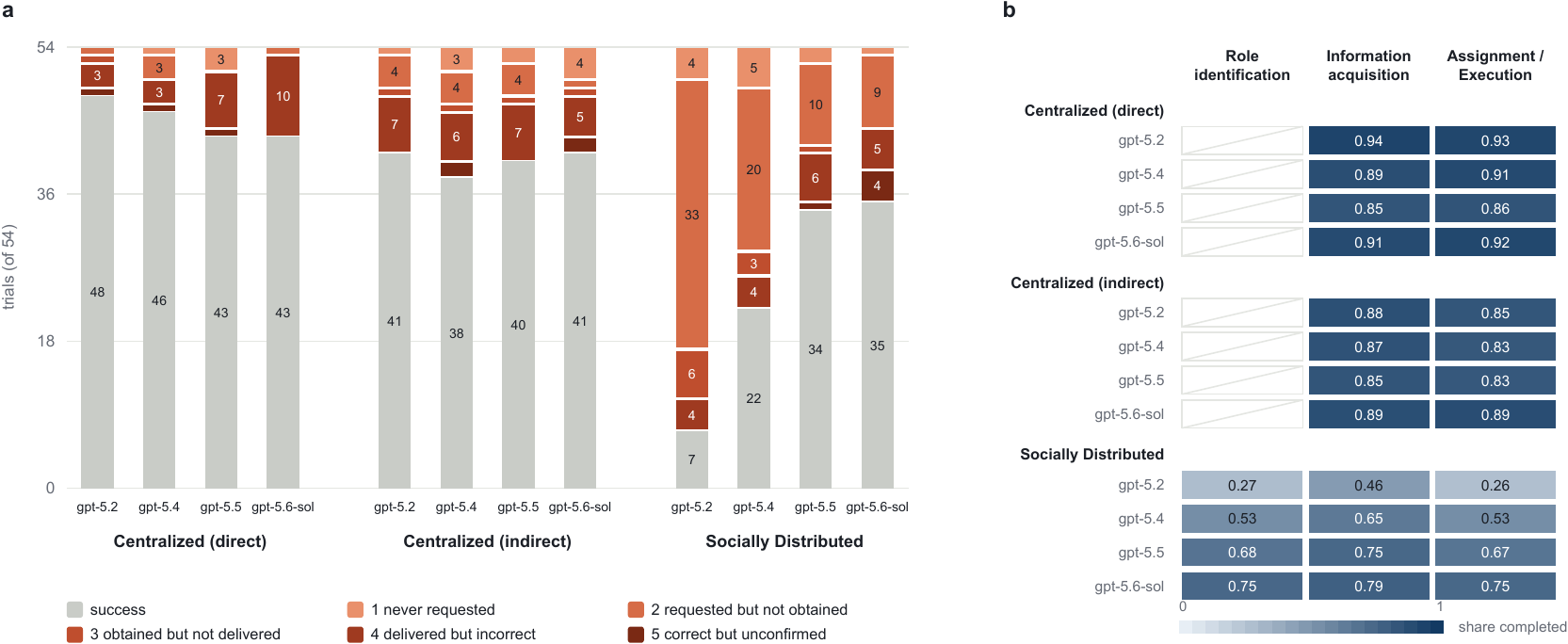}
  \caption{Where trajectories break. (a) Outcome composition of the 54 trials
  per model and condition, with failures stacked in reference-chain order:
  lighter segments mark earlier breaks. (b) Mean process completion against the
  frozen reference; role identification is hatched where the process is absent.}
  \label{fig:failure}
\end{figure}

Figure~\ref{fig:failure} gives the outcome composition of every cell and the
completion of the coordination processes.

\paragraph{The trace labels shift from execution to acquisition.} The three
conditions fail 36, 56 and 118 times out of 216, and their reference-relative
composition differs more than the count. Under both centralized conditions the
largest type is \emph{delivered but incorrect}, 23 of 36 and 25 of 56: needed
reference information was recorded as acquired, but the required action trace
was not correct. Under socially distributed access that type falls to 19 of 118
and the largest becomes \emph{requested but not obtained}, 72 of 118, in which a
needed value remains unrecorded after the agent approached its designated holder.
A further 12 never approached the holder at all. Counting the three types that
attempt no action, 94 of 118 socially distributed failures end without the agent
ever trying to change the state, against 10 of 36 and 27 of 56.

The facts that go missing are service-side. Among the socially distributed trials
of type two, 19 lack only service-side facts and 53 lack both service-side and
customer-side facts, and none lack customer facts alone. The disclosure gate is
not the obstacle. Across both indirect conditions no trial failed because the
agent asked for a fact and the gate withheld it.

\emph{Acquisition-stage labels also covary with the model gaps.} Under socially
distributed access the acquisition failures, types one and two together, fall
from 37 of 54 trials for gpt-5.2 to 25, 12 and 10 for the other three, tracking
success over a range of 27 trials. Execution-stage failures over the same models
are 4, 4, 6 and 5, a range of 2. This descriptive association is consistent with
differences in obtaining needed values, but does not establish whether the
evaluated agent or an error-prone simulated participant caused any failed
acquisition.

\paragraph{The processes agree with the failure counts.} Under socially
distributed access the share of required holders the agent got to act rises with
success, from 0.27 to 0.53, 0.68 and 0.75, and information acquisition rises from
0.46 to 0.79. Merely contacting the right holder is far more common, at 0.62 and
above for every model, so the gap sits in converting contact into action rather
than in finding the participant. In the two centralized conditions every one of
these shares lies between 0.83 and 0.94 for all four models, which is the absent
separability seen one level below the score. Questions reaching a participant
unable to answer them run 5.4 to 7.3 per trial under socially distributed access
against 1.3 to 1.9 under centralized indirect access, the price of having to
discover capability.

\subsection{The cost of reaching through others}

Social access takes 26.2 agent actions and \$1.061 per episode, versus 14.4--13.2
and \$0.252--\$0.467 under centralized access. In the social arm, the weakest
model uses the most actions (33.2) yet has only 0.130 success; the added effort
is primarily messages. Appendix Figure~\ref{fig:costnative} shows both cost scopes.

\section{Related work}
\label{sec:related}

CollabSim varies communication bandwidth, information visibility and group size
in controlled multi-agent experiments~\citep{chen2026collabsim}.
EntCollabBench evaluates a permission-isolated, role-specialized enterprise
organization~\citep{yu2026entcollabbench}, while CollabBench studies cooperative
games with simulated player profiles~\citep{qian2026collabbench}. Incognita
instead builds matched access variants of a fixed source corpus and requires one
evaluated agent to discover service capability.

Work on information arriving across turns studies a single
interlocutor~\citep{laban2026lost,li2025questbench,qian2024tellmemore,wu2025collabllm};
here the agent must decide whom to ask before what to ask. Our reference-relative
taxonomy refines premature termination~\citep{cemri2025mast} without assigning
cause to a participant's response, and we report cost and validity limits
following evaluation-reporting guidance~\citep{kapoor2025agentsthatmatter,zhu2025abc,miller2024errorbars}.

\section{Limitations}
\label{sec:limitations}

\paragraph{Scope and treatment.} One task family, 18 tasks, four one-vendor
models and three repetitions limit generality; task-cell means move in thirds.
Only two of four pre-specified contrasts exclude zero. Separability is post-hoc,
all adjusted interactions involve the earliest snapshot gpt-5.2, and split-half
concordance is inconclusive. The social arm bundles six changes and identifies
no mechanism. Dead-end and contact-to-action gaps motivate a disclosed-
capability-map arm as the next discriminator.

\paragraph{Simulators and measures.} Treatment also changes language-model
simulator error. Without human or desk-fidelity labels, we cannot attribute the
72 \emph{requested but not obtained} failures or test participant error/refusal
across evaluated agents; simulated customers may improvise agreement
outside their cards~\citep{zhou2026sim2real}. The reader is model-free but
reference-relative and not human-validated. A post-hoc check finds
non-reference reads in 12/72; exact-value overlap changes no label
(Appendix~\ref{app:audit}).

\section{Conclusion}
\label{sec:conclusion}

On these eighteen tasks, centralized access separates no model pair while social
access descriptively separates five of six. A post-hoc, multiplicity-adjusted
interaction analysis finds three widened gpt-5.2 gaps. Reference-relative
trajectories associate them with missing acquisition but do not assign cause.
This corpus identifies an access-structure bundle, not its mechanism; simulator
fidelity remains open.

\bibliographystyle{plainnat}
\bibliography{ref}

\appendix

\section{System prompts}
\label{app:prompts}

\newtcolorbox{promptbox}[1]{%
  breakable, enhanced, sharp corners,
  colback=black!3, colframe=black!65, boxrule=0.5pt,
  fonttitle=\bfseries\footnotesize, colbacktitle=black!10, coltitle=black,
  left=4pt, right=4pt, top=3pt, bottom=3pt,
  before skip=8pt, after skip=8pt,
  title={#1}}

\input{appendix_prompts}

\section{Exploratory interaction and corpus-size analyses}
\label{app:interaction}

\subsection{Direct model-by-condition interaction}

This analysis was proposed after the data were seen and is not part of the
pre-specified comparison. For each task $t$ and model pair $(A,B)$, we first
average the three repetitions within each condition and compute
\[
  \Delta_{AB,t}=
  \bigl(S_{\mathrm{indirect},A,t}-S_{\mathrm{social},A,t}\bigr)
  -\bigl(S_{\mathrm{indirect},B,t}-S_{\mathrm{social},B,t}\bigr).
\]
The null treats the two condition labels as exchangeable within a task, which
changes the sign of $\Delta_{AB,t}$. There are $2^{18}$ sign assignments; because
global sign reversal gives the same absolute statistic, we enumerate one member
of each of the $2^{17}$ complementary pairs. We report two-sided exact
sign-flip/permutation $p$ values under this exchangeability assumption. The six
pairwise tests are adjusted by Holm's method;
the intervals are descriptive $t$ intervals for the mean difference-in-
differences, not confidence intervals adjusted for the selection of pairs. The
global and six pairwise statistics were defined together in this post-hoc
analysis; neither appeared in the pre-specified plan.

As a global descriptive test, for every task we subtract the within-task sum of
squared deviations across the four model means under Centralized (indirect)
from that under Socially Distributed. Its task mean is $+0.2222$ and its
two-sided exact sign-flip/permutation $p=0.0127$ under the same exchangeability
assumption. Thus the complete access-structure
bundle changes the dispersion of model scores on this corpus, without
identifying which part of the bundle is responsible.

\begin{table}[h]
  \centering
  \small
  \caption{Exploratory model-by-condition interactions. Positive values mean
  that the first model loses more success under socially distributed access.}
  \label{tab:interaction}
  \begin{tabular}{lrrrr}
    \toprule
    Model pair & Mean $\Delta$ & Descriptive 95\% interval & Exact $p$ & Holm $p$ \\
    \midrule
    gpt-5.2 -- gpt-5.4     & +0.333 & $[+0.128,+0.538]$ & 0.00586 & 0.02344 \\
    gpt-5.2 -- gpt-5.5     & +0.519 & $[+0.276,+0.761]$ & 0.00061 & 0.00366 \\
    gpt-5.2 -- gpt-5.6-sol & +0.519 & $[+0.245,+0.793]$ & 0.00264 & 0.01320 \\
    gpt-5.4 -- gpt-5.5     & +0.185 & $[-0.037,+0.407]$ & 0.13525 & 0.40576 \\
    gpt-5.4 -- gpt-5.6-sol & +0.185 & $[-0.058,+0.428]$ & 0.16895 & 0.40576 \\
    gpt-5.5 -- gpt-5.6-sol & +0.000 & $[-0.273,+0.273]$ & 1.00000 & 1.00000 \\
    \bottomrule
  \end{tabular}
\end{table}

\subsection{Measurement sensitivity audits}
\label{app:audit}

Both checks here were proposed after the final review and use no API calls.
First, 12 of the 72 socially distributed \emph{requested but not obtained}
trials contain 22 accepted non-reference reads. If every missing service-side
item in those 12 trials is credited, regardless of what the read returned, at
most 12 labels change. Five would move to \emph{obtained but not delivered} and
seven to \emph{never requested}; all remain in the acquisition/no-action group,
so the 94/118 count and acquisition--execution asymmetry are unchanged. A
stricter check requires a scalar in the return to match
exactly an argument consumed downstream from the missing reference output. It
finds one such trial, where \texttt{get\_user\_details} returns a needed payment
method, but other items remain missing and no label changes. This check is
deterministic but narrower than semantic equivalence and is not a human audit.

Second, seven tasks carry an operative natural-language assertion channel, for
252 Incognita trials. Ten database passes become failures on that channel: one
under Centralized (direct), four under Centralized (indirect), and five under
Socially Distributed. Thus success is not purely database-deterministic, though
the reference-relative trajectory reader remains model-free. Recomputing the
four pre-specified contrasts from database scores alone leaves all four
zero-crossing decisions unchanged.
The release bundle will include the frozen interpretation table and
coordination reference; no anonymous bundle is available during review.

\subsection{Post-hoc corpus-size extrapolation}

For each condition, we use the median standard deviation of its six observed
task-level pairwise gaps as a plug-in estimate and solve the usual paired-
$t$ approximation for 80\% power at two-sided $\alpha=0.05$ to detect that
condition's largest observed gap. The calculation is post-hoc: it neither
accounts for the six comparisons nor propagates uncertainty in the observed
gap or variance. It is therefore a description of this corpus's observed
resolution, not a prospective power guarantee or economic conclusion. Episode
costs multiply the required task count by four models, three repetitions, and
the recorded per-episode price; they omit development, simulation and offline
evaluation costs.

\begin{table}[h]
  \centering
  \small
  \caption{Post-hoc plug-in corpus sizes required to resolve each arm's largest
  observed model gap. Costs are implied online episode costs only.}
  \label{tab:corpus-size}
  \begin{tabular}{lrrrrr}
    \toprule
    Condition & Gap SD & Largest gap & Tasks & \$/episode & Implied cost \\
    \midrule
    Centralized (direct)   & 0.334 & 0.093 & 104 & 0.252 & \$314 \\
    Centralized (indirect) & 0.312 & 0.056 & 249 & 0.467 & \$1,395 \\
    Socially Distributed   & 0.369 & 0.519 & 6   & 1.061 & \$76 \\
    \bottomrule
  \end{tabular}
\end{table}

\section{Additional figures}
\label{app:figures}

\begin{figure}[h]
  \centering
  \includegraphics[width=\linewidth]{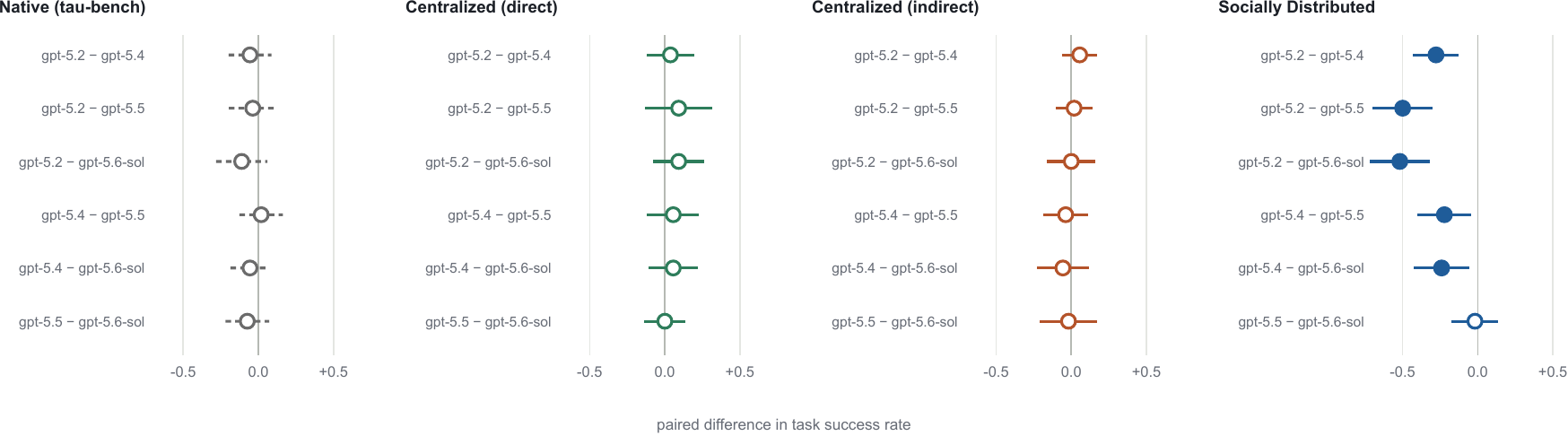}
  \caption{Can the four models be told apart inside each arm? Six
  model-against-model paired differences per arm at the task unit, with
  descriptive $t$ intervals, where a filled dot is an interval that excludes
  zero. Both centralized conditions and the native anchor leave every pair inside
  its interval. Socially Distributed separates five of six. This is a statement
  about measurement resolution and not about which ordering is correct.}
  \label{fig:separability}
\end{figure}

\begin{figure}[h]
  \centering
  \includegraphics[width=\linewidth]{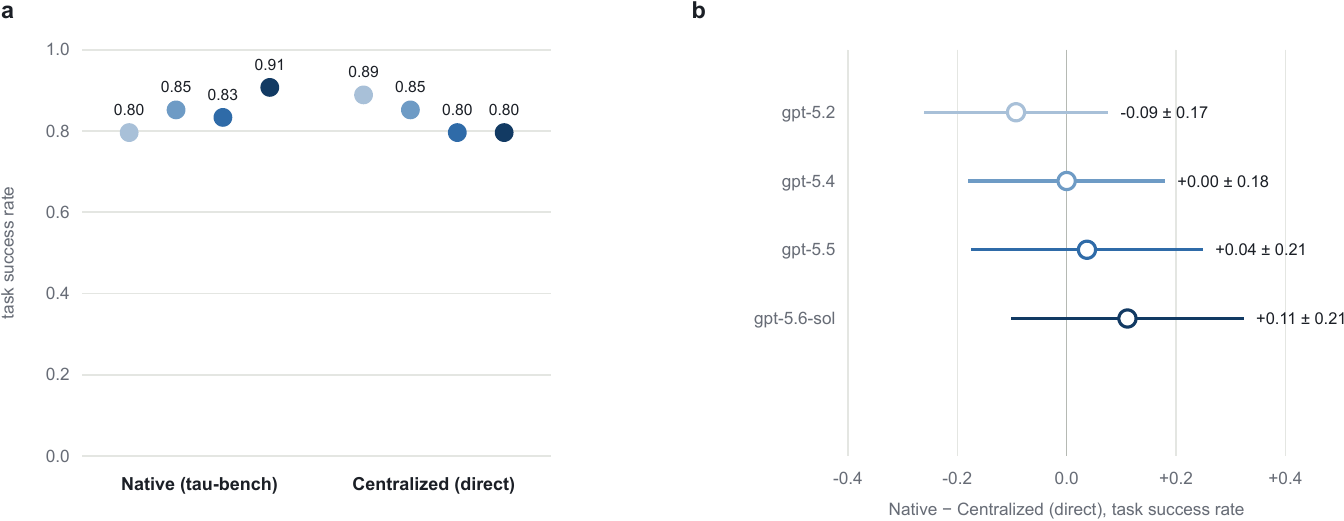}
  \caption{The native $\tau$-bench anchor against Centralized (direct), the
  Incognita condition closest to it. (a) Task success per model. (b) Native minus
  Centralized (direct) at the task unit with a descriptive $t$ interval, where all
  four intervals include zero. The two harnesses differ in several components at
  once, so this comparison bounds the transport gap and supports no causal claim.}
  \label{fig:native}
\end{figure}

\begin{figure}[h]
  \centering
  \includegraphics[width=\linewidth]{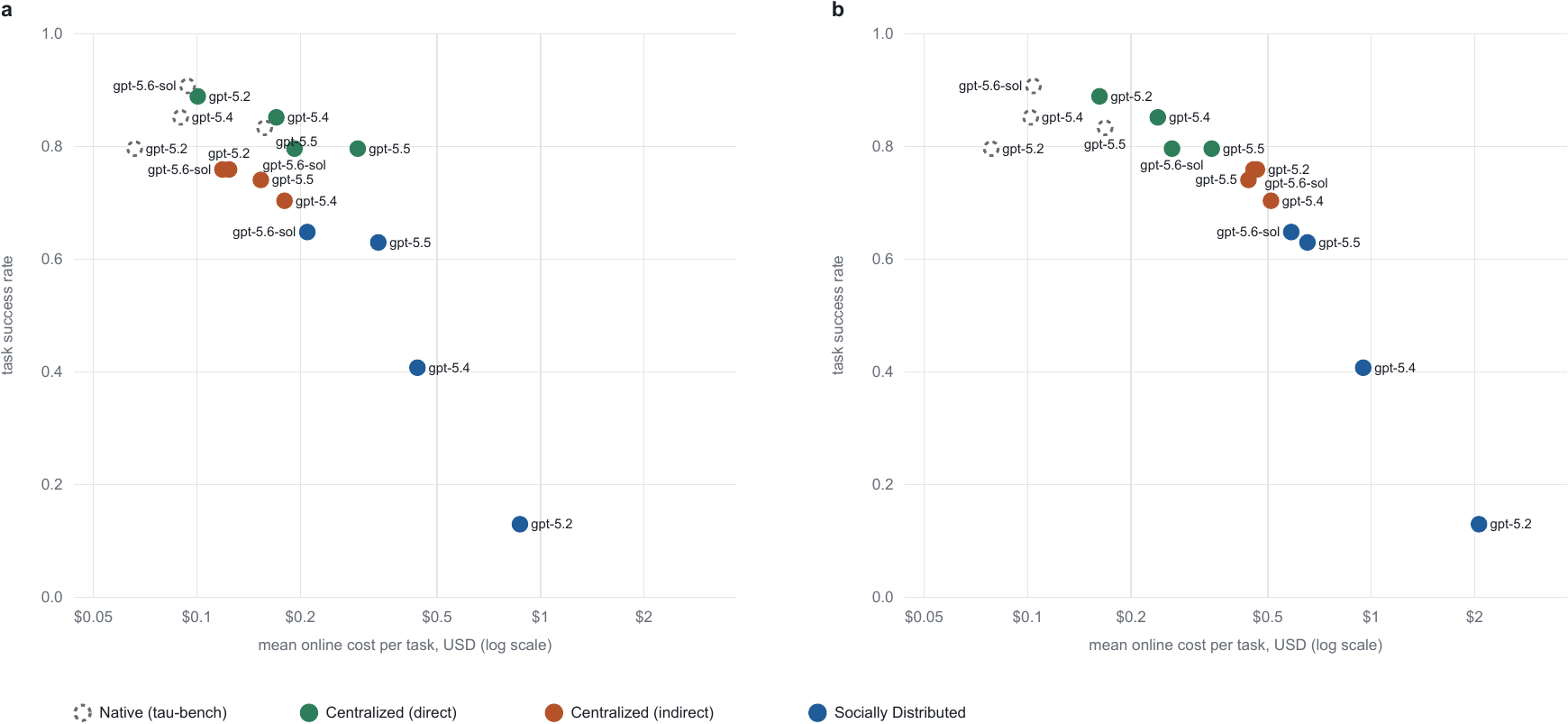}
  \caption{Task success against cost, mean online cost per task on a log scale,
  with the native $\tau$-bench anchor as dashed markers. Colour is the access
  condition and every point carries its model name. (a) The evaluated agent's own
  inference. (b) Every online call the episode needed, including the simulated
  customer and the environment-side participants. The offline judge and evaluator
  are priced in neither scope and both scopes use the common rate table.
  Conditions are never joined and no cost difference is drawn. The anchor runs in
  its own harness and is not on the same cost basis as the three conditions.}
  \label{fig:costnative}
\end{figure}

\begin{figure}[h]
  \centering
  \includegraphics[width=\linewidth]{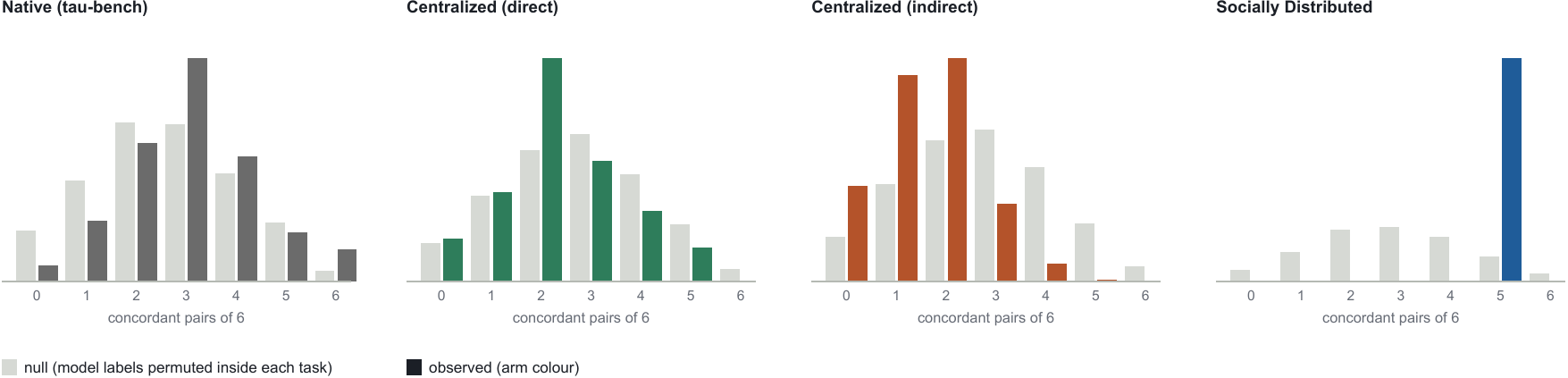}
  \caption{Is each arm's model ordering reproducible across halves of the task
  set? 4000 random balanced splits of the eighteen tasks. Bars show how many of
  the six pairwise model orderings agree between the two halves, in the arm
  colour, against a null that permutes model labels inside each task before
  splitting, in grey. The null keeps every task's four scores and reassigns which
  model owns which, so it preserves task difficulty and breaks only model
  identity. Observed mean concordant pairs are 3.0, 2.3, 1.5 and 5.0, and the
  share of null splits at or above the observed mean is 0.28, 0.56, 0.77 and
  0.15. Exploratory, and not used for the pre-specified comparison.}
  \label{fig:splithalf}
\end{figure}

\begin{figure}[h]
  \centering
  \includegraphics[width=\linewidth]{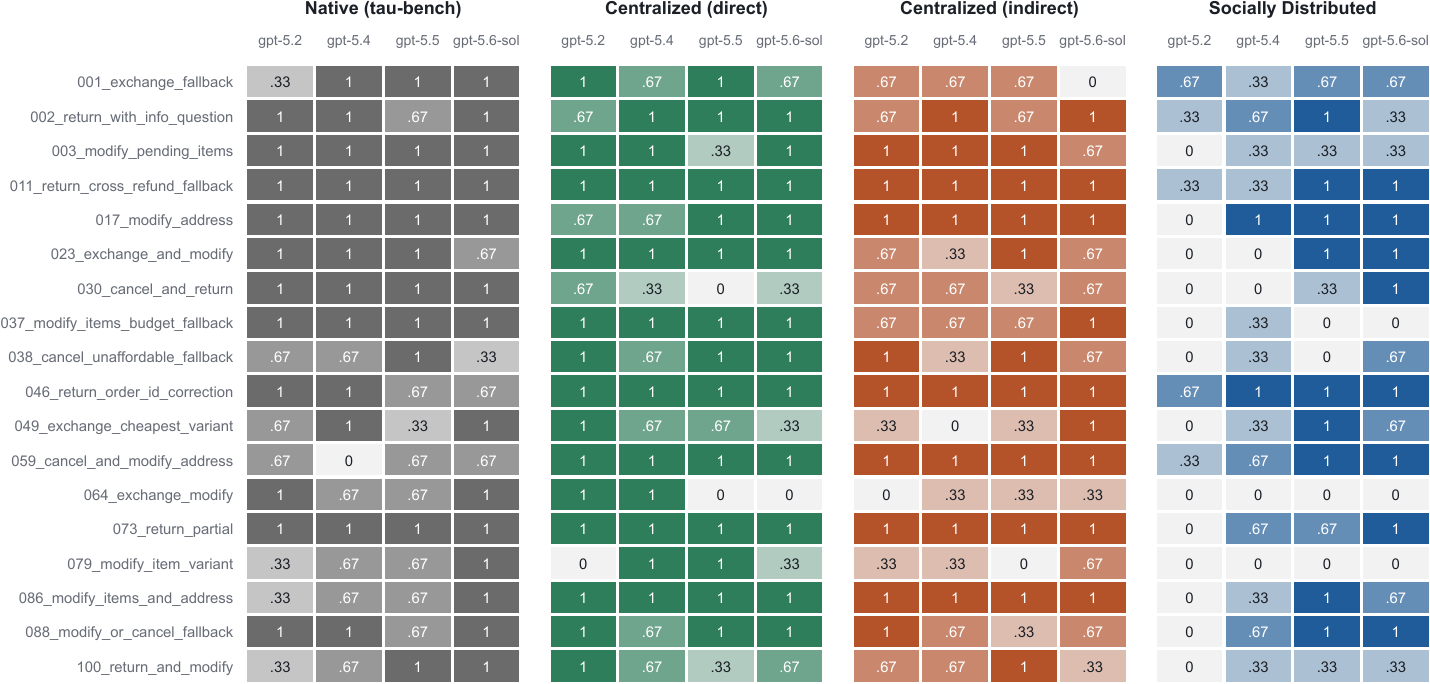}
  \caption{Task success per task, arm and model. Each cell is one task's mean
  over its three repetitions in one arm and model, where darker is higher. Rows
  are the eighteen tasks in identifier order and columns are the four models
  inside each arm. The eight repaired tasks carry their re-run results.}
  \label{fig:heatmap}
\end{figure}

\begin{figure}[h]
  \centering
  \includegraphics[width=\linewidth]{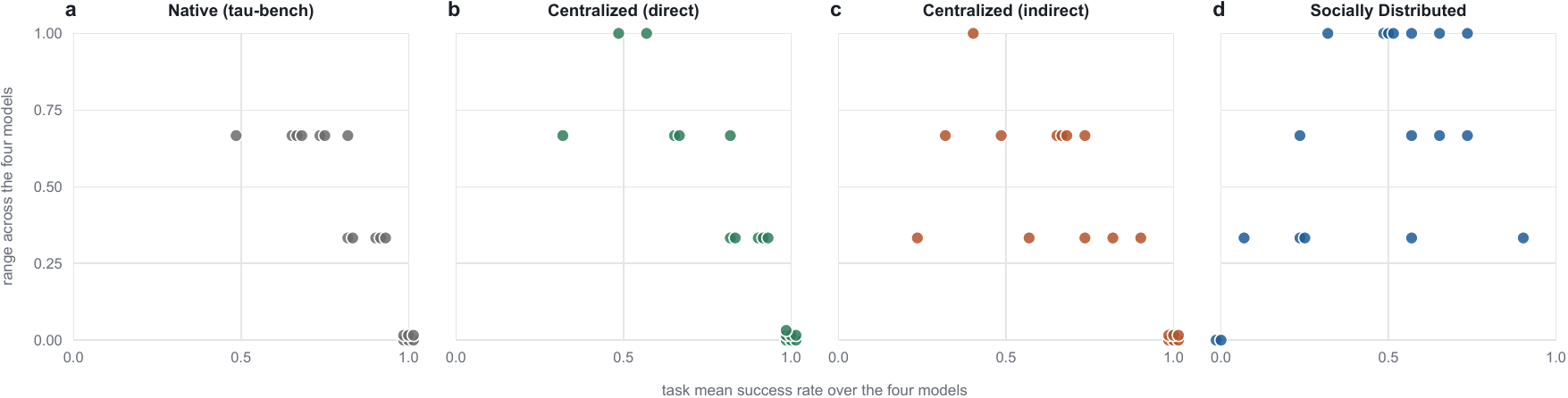}
  \caption{Per-task dispersion across the four models under each access
  condition. Each point is one of the eighteen tasks, positioned horizontally at
  its mean success rate over the four models and vertically at the range between
  the highest and lowest model mean, so the range moves in steps of one third.
  Zero-range tasks in the centralized conditions and the native anchor sit at the
  ceiling. Averaged over tasks the observed range is 0.35, 0.35, 0.37 and 0.63
  for (a) to (d), and a permutation null that reassigns each task's twelve trials
  to the four models at random gives 0.31, 0.28, 0.35 and 0.47, with the share of
  null draws at or above the observed value equal to 0.146, 0.035, 0.422 and
  0.001. This is a descriptive proxy for the information each task carries about
  differences among the models.}
  \label{fig:dispersion}
\end{figure}

\end{document}

%% file: appendix_prompts.tex

This appendix reproduces verbatim the static system prompts of the run the results describe, read out of its trial bundles. The evaluated-agent and service-side prompts are invariant across tasks. The customer prompt has shared behaviour rules and, for eight tasks, a task-specific \texttt{<requests>} block; all eight additions appear below. The customer's task-specific memory is not part of its system prompt, because the disclosure gate releases it one field at a time as per-turn observations. No tool name appears in any prompt, since tool capability is bound separately through the API from the entity-tool whitelist.

\subsection{The evaluated agent}

The agent's prompt carries the participant list, a neutral objective and the output format. It carries no capability list for the other participants, no goal, no policy and no hint about exploration, efficiency or irreversibility. Under Centralized (direct) it also carries the service role and operating rules, because in that condition the agent is the service.

\begin{promptbox}{Evaluated agent, Centralized (direct)}
\begin{Verbatim}[fontsize=\small,breaklines=true,breakanywhere=true]
You are an agent operating in an environment with other participants.

Your objective is to resolve USER's request by interacting with the participants available to you.

<role>
You are the customer-service desk of a retail organization, and the whole of its customer-facing work is yours. From an external detail such as an email address, or a name together with a postal zip code, you find the internal identifier the organization uses for that customer. You look up that customer's profile, an existing order of theirs, and the product catalog, and you update a profile's default shipping address. You carry out the changes available while an order is still in the pending stage, namely cancelling it, or changing its shipping address, its payment method, or the items on it, and you carry out returns and exchanges of items on an order once it has reached the delivered stage.
</role>

<operating_rules>
- A profile holds the customer's name, email, default address, the payment methods on file (gift card, PayPal, or credit card), and the identifiers of their orders.
- The only change you make to a profile is to its default shipping address.
- An order has an identifier, the customer it belongs to, an address, the items on it, a status, fulfillment information, and a payment history.
- An order's status reflects its current lifecycle stage, for example pending, processed, delivered, or cancelled.
- The catalog is organized into product types. Each type has variant items distinguished by their option values, and each item has its own option values, availability, and price.
- A product identifier and an item identifier are unrelated and must not be confused.
- Cancelling an order, or changing its shipping address, its payment method, or the items on it, is available only while the order's status is pending.
- A cancellation needs a reason, and only 'no longer needed' or 'ordered by mistake' are acceptable.
- Changing the payment method replaces it with a single different method, and a gift card used for it must hold enough balance to cover the order total.
- An item can be changed or exchanged only to a different available option of the same product type. Any price difference is settled through a payment method the requester provides, and a gift card used for it must hold enough balance.
- The items on an order can be changed only once, and after that the order can no longer be modified or cancelled.
- Returning or exchanging items on an order is available only once the order's status is delivered.
- A return sends the refund either to the original payment method or to an existing gift card.
- A return or an exchange can be done only once on an order.
- Before you carry out any change, you restate its details and proceed only once the requester confirms them.
</operating_rules>

Each turn, take exactly one action: call exactly one tool, or send one message to USER. Sending a message never ends the episode.

The episode ends only when you end it, by sending message_type FINAL to ENVIRONMENT. ENVIRONMENT is not one of your recipients and never replies. It only receives this signal. After you send it you receive no further turns, so send your closing response to USER before you end the episode.

To send a message or to end the episode, output one JSON object:
- To send a message: {"recipient": "USER", "message": "<text>", "message_type": "MESSAGE"}
- To end the episode: {"recipient": "ENVIRONMENT", "message": "done", "message_type": "FINAL"}
Every turn is either exactly one tool call with no text alongside it, or exactly one JSON object with nothing else. Never both in the same turn.
\end{Verbatim}
\end{promptbox}

\begin{promptbox}{Evaluated agent, Centralized (indirect)}
\begin{Verbatim}[fontsize=\small,breaklines=true,breakanywhere=true]
You are an agent operating in an environment with other participants.

Your objective is to resolve USER's request by interacting with the participants available to you.

Each turn, send one message to exactly one recipient: USER, Vega. Sending a message never ends the episode.

The episode ends only when you end it, by sending message_type FINAL to ENVIRONMENT. ENVIRONMENT is not one of your recipients and never replies. It only receives this signal. After you send it you receive no further turns, so send your closing response to USER before you end the episode.

Output one JSON object per turn:
- To send a message: {"recipient": "<USER | Vega>", "message": "<text>", "message_type": "MESSAGE"}
- To end the episode: {"recipient": "ENVIRONMENT", "message": "done", "message_type": "FINAL"}
Output only the JSON object, nothing else.
\end{Verbatim}
\end{promptbox}

\begin{promptbox}{Evaluated agent, Socially Distributed}
\begin{Verbatim}[fontsize=\small,breaklines=true,breakanywhere=true]
You are an agent operating in an environment with other participants.

Your objective is to resolve USER's request by interacting with the participants available to you.

Each turn, send one message to exactly one recipient: USER, Aster, Boreal, Cygnus, Dorian, Elara, Fenn. Sending a message never ends the episode.

The episode ends only when you end it, by sending message_type FINAL to ENVIRONMENT. ENVIRONMENT is not one of your recipients and never replies. It only receives this signal. After you send it you receive no further turns, so send your closing response to USER before you end the episode.

Output one JSON object per turn:
- To send a message: {"recipient": "<USER | Aster | Boreal | Cygnus | Dorian | Elara | Fenn>", "message": "<text>", "message_type": "MESSAGE"}
- To end the episode: {"recipient": "ENVIRONMENT", "message": "done", "message_type": "FINAL"}
Output only the JSON object, nothing else.
\end{Verbatim}
\end{promptbox}

\subsection{The customer simulator}

The shared behaviour block is identical in every trial. Eight of the eighteen tasks append a task-specific request block. Each complete USER prompt is identical across access conditions, evaluated models and repetitions for the same task.

\begin{promptbox}{USER, shared behaviour rules}
\begin{Verbatim}[fontsize=\small,breaklines=true,breakanywhere=true]
You are a customer with a request to resolve.
You communicate only by sending messages to the agent, the way a real customer naturally would, in plain sentences with no lists or formatting.
Use only your <memory> to answer, and if something you are asked for is not in it, say you are not sure.
Never invent, assume, or settle for a fact or preference your <memory> does not state, and never confirm one the agent states for you unless your <memory> states it too.
\end{Verbatim}
\end{promptbox}

\subsubsection{Task-specific customer request blocks}

Only the following eight tasks append text to the shared behaviour rules. The blocks are reproduced in full.

\begin{promptbox}{USER requests, \texttt{retail\_002\_return\_with\_info\_question}}
\begin{Verbatim}[fontsize=\small,breaklines=true,breakanywhere=true]
<requests>
Besides answering the agent, you have your own requests to raise. Bring each of these up yourself during the conversation, in your own words; they are your own demands, not answers to the agent.
- You want the agent to tell you how many t-shirt options are currently available in the online store.
</requests>
\end{Verbatim}
\end{promptbox}

\begin{promptbox}{USER requests, \texttt{retail\_003\_modify\_pending\_items}}
\begin{Verbatim}[fontsize=\small,breaklines=true,breakanywhere=true]
<requests>
Besides answering the agent, you have your own requests to raise. Bring each of these up yourself during the conversation, in your own words; they are your own demands, not answers to the agent.
- You want the agent to tell you how many t-shirt options the store currently offers.
</requests>
\end{Verbatim}
\end{promptbox}

\begin{promptbox}{USER requests, \texttt{retail\_011\_return\_cross\_refund\_fallback}}
\begin{Verbatim}[fontsize=\small,breaklines=true,breakanywhere=true]
<requests>
Besides answering the agent, you have your own requests to raise. Bring each of these up yourself during the conversation, in your own words; they are your own demands, not answers to the agent.
- You have two payment methods and two orders, and you want each order refunded to the other order's payment method.
- If the agent says that is not possible, you are angry and swear a few times, then agree to return everything with each order's original payment method.
</requests>
\end{Verbatim}
\end{promptbox}

\begin{promptbox}{USER requests, \texttt{retail\_030\_cancel\_and\_return}}
\begin{Verbatim}[fontsize=\small,breaklines=true,breakanywhere=true]
<requests>
Besides answering the agent, you have your own requests to raise. Bring each of these up yourself during the conversation, in your own words; they are your own demands, not answers to the agent.
- Before anything else, you want the agent to tell you the tracking number of the order that contains the tablet.
</requests>
\end{Verbatim}
\end{promptbox}

\begin{promptbox}{USER requests, \texttt{retail\_037\_modify\_items\_budget\_fallback}}
\begin{Verbatim}[fontsize=\small,breaklines=true,breakanywhere=true]
<requests>
Besides answering the agent, you have your own requests to raise. Bring each of these up yourself during the conversation, in your own words; they are your own demands, not answers to the agent.
- First, you want to know whether the agent can split the payment for the order with another card.
- Only if the agent says that is not possible: you want the agent to tell you which single item in the order is the most expensive and how much it costs, and whether you can just cancel that one item.
- Only if that is not possible either: you ask whether all the items can be switched to their cheapest options to bring the total down to what your card can cover. If that is possible, you confirm and ask the agent to do it.
- Only if that is not possible either: you ask the agent to just cancel the order so that you can order again.
- You do not end the conversation until your changes have been made.
</requests>
\end{Verbatim}
\end{promptbox}

\begin{promptbox}{USER requests, \texttt{retail\_038\_cancel\_unaffordable\_fallback}}
\begin{Verbatim}[fontsize=\small,breaklines=true,breakanywhere=true]
<requests>
Besides answering the agent, you have your own requests to raise. Bring each of these up yourself during the conversation, in your own words; they are your own demands, not answers to the agent.
- First, you want to know whether the agent can split the payment for the order with another card.
- Only if the agent says that is not possible: you want the agent to tell you which item in the order is the most expensive and how much it costs, and whether you can just cancel that one item.
- Only if that is not possible either: you want to know whether all the items can be switched to their cheapest options to bring the total down to what your card can cover.
- Only if that is not possible either: you ask the agent to just cancel the order, since you no longer need this one and will place a new one instead.
</requests>
\end{Verbatim}
\end{promptbox}

\begin{promptbox}{USER requests, \texttt{retail\_046\_return\_order\_id\_correction}}
\begin{Verbatim}[fontsize=\small,breaklines=true,breakanywhere=true]
<requests>
Besides answering the agent, you have your own requests to raise. Bring each of these up yourself during the conversation, in your own words; they are your own demands, not answers to the agent.
- You ask the agent explicitly to provide the refund within 3 days.
- You also want the agent to tell you the total amount of the refund you should expect.
- After the return is complete, you want the agent to tell you the total amount you paid for the remaining items on the same order.
</requests>
\end{Verbatim}
\end{promptbox}

\begin{promptbox}{USER requests, \texttt{retail\_059\_cancel\_and\_modify\_address}}
\begin{Verbatim}[fontsize=\small,breaklines=true,breakanywhere=true]
<requests>
Besides answering the agent, you have your own requests to raise. Bring each of these up yourself during the conversation, in your own words; they are your own demands, not answers to the agent.
- First, you want to know why your two orders, #W2702727 and #W8268610, are both still pending when #W8268610 was placed much earlier in the year.
- You find the wait on the older order, #W8268610, unreasonable. If the agent cannot guarantee that it will be processed within 5 days, you want it cancelled, and you want the agent to tell you the total refund amount.
- For order #W2702727, you want the shipping address switched to your new home in a different city, because you will move before it is delivered next month. You want the agent to confirm the change, confirm the order will be delivered to the new address, and tell you the total price of the order after the change.
</requests>
\end{Verbatim}
\end{promptbox}

\subsection{Service-side participants}

Under Centralized (indirect) a single participant holds every role and every tool. Under Socially Distributed the same roles and tools are partitioned across six entities. Each entity prompt is its own \texttt{<role>} block, then the shared \texttt{<conduct>} block reproduced once below, then its own \texttt{<operating\_rules>} block. Elara is read-only and has no operating rules.

\begin{promptbox}{Shared \texttt{<conduct>} block, identical for Vega and all six entities}
\begin{Verbatim}[fontsize=\small,breaklines=true,breakanywhere=true]
<conduct>
1. Reactive disclosure: answer what you are asked rather than volunteering an overview of everything you handle.
2. Precision mirroring: match the precision of your answer to the precision of the request, so a broad question gets one neutral sentence about your work and a request too vague to act on gets a question for the single missing detail.
3. Grounded answers: state only what your records actually show, and if a requested detail is not there, say so rather than inventing or guessing it.
4. Remit boundary: handle requests that fall within your remit, and decline anything outside it plainly.
5. Non-referral: never redirect the asker to someone else, name someone else, or answer on someone else's behalf.
6. Plain wording: write in plain sentences only, with no bullet points, headings, bold, or other formatting.
7. Findings discipline: when you look something up, give a brief summary rather than everything you find, let the asker draw out the specifics they need, and never name the tools or methods behind it.
</conduct>
\end{Verbatim}
\end{promptbox}

\begin{promptbox}{Vega, Centralized (indirect), all roles and every tool}
\begin{Verbatim}[fontsize=\small,breaklines=true,breakanywhere=true]
<role>
You are the customer-service desk of a retail organization, and the whole of its customer-facing work is yours. From an external detail such as an email address, or a name together with a postal zip code, you find the internal identifier the organization uses for that customer. You look up that customer's profile, an existing order of theirs, and the product catalog, and you update a profile's default shipping address. You carry out the changes available while an order is still in the pending stage, namely cancelling it, or changing its shipping address, its payment method, or the items on it, and you carry out returns and exchanges of items on an order once it has reached the delivered stage.
</role>

<operating_rules>
- A profile holds the customer's name, email, default address, the payment methods on file (gift card, PayPal, or credit card), and the identifiers of their orders.
- The only change you make to a profile is to its default shipping address.
- An order has an identifier, the customer it belongs to, an address, the items on it, a status, fulfillment information, and a payment history.
- An order's status reflects its current lifecycle stage, for example pending, processed, delivered, or cancelled.
- The catalog is organized into product types. Each type has variant items distinguished by their option values, and each item has its own option values, availability, and price.
- A product identifier and an item identifier are unrelated and must not be confused.
- Cancelling an order, or changing its shipping address, its payment method, or the items on it, is available only while the order's status is pending.
- A cancellation needs a reason, and only 'no longer needed' or 'ordered by mistake' are acceptable.
- Changing the payment method replaces it with a single different method, and a gift card used for it must hold enough balance to cover the order total.
- An item can be changed or exchanged only to a different available option of the same product type. Any price difference is settled through a payment method the requester provides, and a gift card used for it must hold enough balance.
- The items on an order can be changed only once, and after that the order can no longer be modified or cancelled.
- Returning or exchanging items on an order is available only once the order's status is delivered.
- A return sends the refund either to the original payment method or to an existing gift card.
- A return or an exchange can be done only once on an order.
- Before you carry out any change, you restate its details and proceed only once the requester confirms them.
</operating_rules>
\end{Verbatim}
\end{promptbox}

\begin{promptbox}{Aster, Socially Distributed}
\begin{Verbatim}[fontsize=\small,breaklines=true,breakanywhere=true]
<role>
You are the customer-account desk, one of several specialized desks in a retail customer-service organization. You work with the account-level information on a customer's profile, namely their name, email, default shipping address, the payment methods on file, and the identifiers of their orders. You look that profile up, and you make one change to it, updating the default shipping address.
</role>

<operating_rules>
- A profile holds the customer's name, email, default address, the payment methods on file (gift card, PayPal, or credit card), and the identifiers of their orders.
- The only change you make is to the default shipping address.
- Before you change the address, you restate the new address and proceed only once the requester confirms it.
</operating_rules>
\end{Verbatim}
\end{promptbox}

\begin{promptbox}{Boreal, Socially Distributed}
\begin{Verbatim}[fontsize=\small,breaklines=true,breakanywhere=true]
<role>
You are the returns-and-exchanges desk, one of several specialized desks in a retail customer-service organization. You carry out returns and exchanges of items on an order once it has reached the delivered stage.
</role>

<operating_rules>
- You act only on an order whose status is delivered.
- An exchange changes an item to a different available option of the same product type. Any price difference is settled through a payment method the requester provides, and a gift card used for it must hold enough balance.
- A return sends the refund either to the original payment method or to an existing gift card.
- A return or an exchange can be done only once on an order.
- Before you carry out a return or an exchange, you restate its details and proceed only once the requester confirms them.
</operating_rules>
\end{Verbatim}
\end{promptbox}

\begin{promptbox}{Cygnus, Socially Distributed}
\begin{Verbatim}[fontsize=\small,breaklines=true,breakanywhere=true]
<role>
You are the product-catalog desk, one of several specialized desks in a retail customer-service organization. You provide read-only access to the catalog, at the level of product types, the variant items each type offers, and each item's option values, availability, and price. Catalog inspection is the whole of your work.
</role>

<operating_rules>
- The catalog is organized into product types. Each type has variant items distinguished by their option values, and each item has its own option values, availability, and price.
- A product identifier and an item identifier are unrelated and must not be confused.
</operating_rules>
\end{Verbatim}
\end{promptbox}

\begin{promptbox}{Dorian, Socially Distributed}
\begin{Verbatim}[fontsize=\small,breaklines=true,breakanywhere=true]
<role>
You are the order-records desk, one of several specialized desks in a retail customer-service organization. You provide read-only access to an existing order, so that its current status, its contents, its shipping address, its fulfillment information, and its payment history can be looked up. Inspection is the whole of your work.
</role>

<operating_rules>
- An order has an identifier, the customer it belongs to, an address, the items on it, a status, fulfillment information, and a payment history.
- An order's status reflects its current lifecycle stage, for example pending, processed, delivered, or cancelled.
</operating_rules>
\end{Verbatim}
\end{promptbox}

\begin{promptbox}{Elara, Socially Distributed}
\begin{Verbatim}[fontsize=\small,breaklines=true,breakanywhere=true]
<role>
You are the identity-lookup desk, one of several specialized desks in a retail customer-service organization. From an external detail such as an email address, or a name together with a postal zip code, you find the internal identifier the organization uses for that customer. This single lookup is the whole of your work.
</role>
\end{Verbatim}
\end{promptbox}

\begin{promptbox}{Fenn, Socially Distributed}
\begin{Verbatim}[fontsize=\small,breaklines=true,breakanywhere=true]
<role>
You are the pending-order-operations desk, one of several specialized desks in a retail customer-service organization. You carry out the changes available while an order is still in the pending stage, namely cancelling it, or changing its shipping address, its payment method, or the items on it.
</role>

<operating_rules>
- You act only on an order whose status is pending.
- A cancellation needs a reason, and only 'no longer needed' or 'ordered by mistake' are acceptable.
- Changing the payment method replaces it with a single different method, and a gift card used for it must hold enough balance to cover the order total.
- An item can be changed only to a different available option of the same product type. Any price difference is settled through a payment method the requester provides, and a gift card used for it must hold enough balance.
- The items on an order can be changed only once, and after that the order can no longer be modified or cancelled.
- Before you carry out any change, you restate its details and proceed only once the requester confirms them.
</operating_rules>
\end{Verbatim}
\end{promptbox}